\documentclass[conf]{new-aiaa}
\usepackage[utf8]{inputenc}

\usepackage{graphicx}
\usepackage{amsmath}
\usepackage[version=4]{mhchem}
\usepackage{siunitx}
\usepackage{longtable,tabularx}
\usepackage{algorithm}
\usepackage{algpseudocode}
\title{Monte Carlo Tree Search Based Tactical Maneuvering}

\author{Kunal Srivastava \footnote{Staff Research Engineer, Autonomous Intelligent Systems Department, kunal.srivastava@rtx.com.} and Amit Surana \footnote{Technical Fellow, Autonomous Intelligent Systems Department, amit.surana@rtx.com}}
\affil{Raytheon Technologies Research Center, East Hartford, CT, 06111}

\algtext*{EndWhile}
\algtext*{EndIf}
\algtext*{EndProcedure}
\DeclareMathOperator*{\argmax}{arg\,max}

\newcommand{\tnode}{v}
\newcommand{\pSet}{\mathbb{N}}
\newcommand{\cSet}{\mathbb{S}}
\newcommand{\cSetT}{\mathbb{Z}}
\newcommand{\cSetD}{\mathbb{D}}

\newcommand{\actg}{a}
\newcommand{\actp}{a_1}
\newcommand{\acto}{a_2}
\newcommand{\act}{\actp,\acto}
\newcommand{\actSet}{\mathbb{A}}
\newcommand{\trans}{T}
\newcommand{\Qval}{Q}
\newcommand{\selPol}{\pi}
\newcommand{\stat}{\mathbf{t}}

\newcommand{\payoffelem}{P}
\newcommand{\payoff}{\mathbf{\payoffelem}}
\newcommand{\probelem}{u}
\newcommand{\probvec}{\mathbf{\probelem}}

\newcommand{\R}{\mathbb{R}}

\newcommand{\state}{x}
\newcommand{\statevec}{\mathbf{\state}}
\newcommand{\ctr}{u}
\newcommand{\ctrvec}{\mathbf{\ctr}}
\newcommand{\param}{p}
\newcommand{\paramvec}{\mathbf{\param}}

\newcommand{\pos}{\mathbf{r}}
\newcommand{\velvec}{\mathbf{v}}
\newcommand{\att}{\mathbf{\theta}}
\newcommand{\omgvec}{\mathbf{\omega}}

\newcommand{\ctrlset}{\mathbb{U}}
\newcommand{\ns}{n}
\newcommand{\nc}{m}

\newcommand{\dyn}{\mathbf{f}}

\newcommand{\cstate}{s}
\newcommand{\cstatevec}{\mathbf{\cstate}}

\newcommand{\rstate}{c}
\newcommand{\rstatevec}{\mathbf{\rstate}}

\newcommand{\brg}{\psi}
\newcommand{\angoff}{\phi}
\newcommand{\dist}{d}
\newcommand{\distvec}{\mathbf{\dist}}
\newcommand{\utl}{R}

\newcommand{\disfac}{\gamma}
\newcommand{\prob}{P}
\newcommand{\pol}{\pi}

\newcommand{\thor}{K}

\DeclareGraphicsExtensions{.pdf,.jpeg,.png,.jpg,.emf}

\begin{document}

\maketitle

\begin{abstract}
In this paper we explore the application of simultaneous move Monte Carlo Tree Search
(MCTS) based online framework for tactical maneuvering between two unmanned aircrafts.
Compared to other techniques, MCTS enables efficient search over long horizons and uses
self-play to select best maneuver in the current state while accounting for the opponent aircraft
tactics. We explore different algorithmic choices in MCTS and demonstrate the framework
numerically in a simulated 2D tactical maneuvering application.
\end{abstract}

\section{Nomenclature}

{\renewcommand\arraystretch{1.0}
\noindent\begin{longtable*}{@{}l @{\quad=\quad} l@{}}
$a$  & aircraft action\\
$\actSet$  & aircraft action set\\
$\Delta t $  & time step \\
$ \cSetD$  & set of decision game states\\
$ \dyn$  & aircraft dynamic model \\
$ k$  & discrete time step \\
$\thor_{\mbox{game}}$ & total time game is simulated \\
$\thor_{\mbox{sim}}$ & roll out time in simulation stage of MCTS \\
$M_{\mbox{tree}}$ & maximum allowed MCTS tree size\\
$ N(\tnode)$  & number of times tree node $\tnode$ has been visited \\
$ \pol$  & selection policy in MCTS\\
$ \payoff$  & payoff matrix\\
$ \Qval(\tnode)$  & total reward accumulated at node $\tnode$ \\
$ \pSet$  & set of players \\
$ \pos$  & aircraft position vector \\
$ \utl$  & reward function\\
$ \cstatevec$  & game state vector \\
$\cSet$ & game state space\\
$ \stat(\tnode)$  & statistics collected at node $\tnode$ \\
$ \ctrvec$  & control input vector for aircraft model\\
$ \tnode$  & MCTS tree node \\
$ \velvec$  & aircraft velocity vector \\
$ \statevec$ & aircraft state vector  \\
$ \cSetT$  & set of terminal game states
\end{longtable*}}

\section{Introduction}
In this paper we explore the application of Monte Carlo Tree Search (MCTS) based online framework for tactical
maneuvering between two unmanned aircrafts. The ability of unmanned aircraft to autonomously synthesize and execute
agile maneuvers (e.g. evasive or attack patterns) in complex and dynamic environments is an enabling technology
for future air mission scenarios driven by performance and safety goals \cite{ure2012autonomous}. The tactical maneuvering problem and its several variants such as pursuit evasion game has been studied in the literature. A rule-based adaptive approach was
developed in \cite{burgin1988rule} but requires hard coding of maneuvering logic which can become intractable. Authors in \cite{austin1990game,acematrixgame} suggest
using a game theoretic approach involving a recursive search over discrete maneuver choices to maximize a heuristic
scoring function with a fixed planning horizon. Along similar lines, \cite{virtanen2006modeling} computed a feedback Nash equilibrium of the
dynamic game at each decision stage. However, to keep the computation tractable only a limited planning horizon was
considered. Although a limited planning horizon can mitigate the computational complexity, long planning horizons are
essential to making good maneuver choices. A real-time game theoretic evasion controller based on nonlinear model
predictive approach was proposed in \cite{eklund2005implementing}. The authors commented on the need to encode proven aircraft maneuvering
tactics into the cost functions because the method, by itself, did not produce the required behaviors.

Mcgrew et al. \cite{mcgrew2010air} regarded the pursuit evasion game as a dynamic programming problem, and to avoid the curse of dimensionality used an approximate dynamic programming (ADP) approach for learning a maneuver selection policy. ADP approach provides a fast response to a rapidly changing tactical situation, long planning horizons, and good performance without explicit coding of maneuvering tactics. Along similar lines reinforcement learning based approaches have also been explored in \cite{zhangresearch,yang2019maneuver,bertram2019efficient}. All these methods, however need to assume tactics of the opponent aircraft for learning the policy and thus the performance is limited by that choice.

In order to address the challenge of long term planning and being able to reduce reliance on assumed opponent behavior, we propose to use MCTS. MCTS being a highly selective best first search quickly focuses on promising part of the game tree enabling search over longer horizons without significantly increasing computational burden. Moreover, MCTS uses self-play and thus does not rely on a model of opponent tactics. MCTS based game playing has recently shown human level performance in highly complex games such as Go \cite{silver2016mastering}. MCTS has also recently been applied in urban air mobility scenario \cite{yang2020scalable}. In our application we use simultaneous move MCTS (SMCTS) since both aircrafts simultaneously take the maneuvering decisions to gain position of tactical advantage. Following previous studies mentioned above, we consider a game theoretic representation with actions/moves restricted to a set of discrete maneuvers for each aircraft and assume a perfect information setting i.e. each aircraft is assumed to have exact knowledge of states of the other aircraft at every instant of time. We present different variations of the SMCTS algorithm for the maneuvering problem, and show that it can outperform a short horizon matrix game approach as proposed in \cite{acematrixgame}.

The paper is organized into five sections. Section \ref{sec:bckg} provides the background on the MCTS and SMCTS approaches. The maneuvering problem is discussed in section \ref{sec:prob} along with its game theoretic representation. Different variations of SMCTS approach which we explore in this paper are summarized in section \ref{sec:algo}. Preliminary simulation results are provided in section \ref{sec:sim} along with a discussion on the expected results for the final paper. Finally, section \ref{sec:conc} lists some directions for future research.

\section{Background}\label{sec:bckg}
Consider a two person game described by a tuple $\{\pSet,\cSet=\cSetD\bigcup\cSetT,\actSet,\utl_1,\utl_2,\cstatevec_0\}$.
The player set $\pSet =\{1,2\}$ contains player labels, and by convention a player
is denoted by $i\in \pSet$. $\cSet$ is a set of states, with $\cSetT$ denoting the
terminal states where the game ends and $\cSetD$ the states where players make decisions.
$\actSet = \actSet_1 \times \actSet_2$ is the set of joint moves of individual players.
We denote $\actSet_i(\cstatevec)$ the moves available to player $i$ in the state $\cstatevec \in \cSet$.
The transition function $\trans: \cSet \times \actSet_1 \times \actSet_2 \rightarrow \cSet$ defines the successor state given a current state and moves of both the players. The reward functions $\utl_i:\cSet\rightarrow \R$ gives the reward/payoff of player $i$. The game begins in an initial state $\cstatevec_0$. We are interested in online approaches for solving the game.

\subsection{Monte Carlo Tree Search}
\begin{figure}
  \begin{center}
  \includegraphics[trim=0cm 11cm 0cm 0cm,clip,scale=0.55]{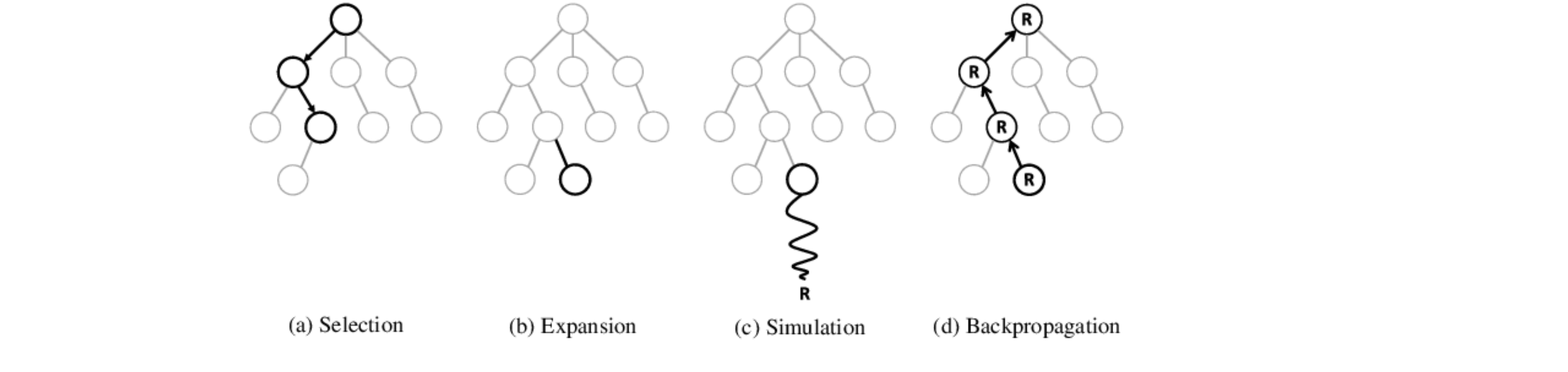}
  \caption{Schematic of the four steps of MCTS}\label{fig:mcts}
  \end{center}
\end{figure}

Monte-Carlo Tree Search (MCTS) is a highly selective best first search that relies on random simulations to estimate state values in a game tree \cite{browne2012survey}. The central data structure in MCTS is the game tree in which nodes correspond to game states and edges correspond to possible actions or moves. The role of this tree is two-fold: it stores the outcomes of random simulations and it is used to bias random simulations towards promising sequences of moves. MCTS is divided in four main steps (see Fig. ~\ref{fig:mcts})that are repeated until a prescribed computational budget is met:
\begin{itemize}
  \item \textit{Selection}: This step aims at selecting a node in the tree from which a new random simulation will be performed.  The most popular selection strategy is upper confidence bound for trees (UCT) \cite{kocsis2006bandit}, which from a node $\tnode$ selects the child $\tnode^\prime$ with the highest score, i.e.
      \begin{equation}\label{eq:uct}
      \argmax_{\tnode^\prime \mbox{chidren of} \tnode} \frac{\Qval(\tnode^\prime)}{N(\tnode^\prime)}+c\sqrt{\frac{2N(\tnode)}{N(\tnode^\prime)}}.
      \end{equation}
      In this formula, the first term denotes the average score, i.e., the win rate, of node $\tnode^\prime$ with $\Qval(\tnode)$ representing the total reward accumulated at node $\tnode$, $N(\tnode^\prime)$ and $N(\tnode)$ denote the total number of times child $\tnode^\prime$ and its parent $\tnode$ have been visited, respectively, and $c$ is a constant, which balances exploration vs. exploitation. This selection strategy is applied until a node is reached that is not fully expanded, i.e., not all of its children have been added to the tree yet.
  \item \textit{Expansion}: If the selected node does not end the game, this steps expands and adds a new leaf node to the selected one.
  \item \textit{Simulation}: This step starts simulating the game with self-play from the state associated with the selected leaf node while executing random moves until the end of the game is reached, and returns the game reward. Note that during simulation step MCTS only requires a black box simulator, and can thus be applied in problems that are too large or too complex to represent with explicit probability distributions. It uses random moves during simulations to estimate the potential for long-term reward, and is often effective without any search heuristics or prior domain knowledge. However, game knowledge can be incorporated in form of a playout strategy to make the playouts more realistic. One approach to improve the quality of the playouts is by applying $\epsilon$-greedy playouts. For each move played in the playouts, there is a probability $\epsilon$ that a random move is played. Otherwise, domain knowledge can be used to assign a value to each valid move for the current player. The move with the highest heuristic value is played.
  \item \textit{Backpropagation}: In this step the result of the playout is propagated back along the previously traversed path up to the root node. The most popular and also the most effective backpropagation strategy is Average, which keeps track of the average of the results of all playouts through each node. Other strategies include Max, Informed Average, and Mix \cite{coulom2006efficient}.
\end{itemize}
These four phases are repeated either a fixed number of times or until the time runs out. After the search is finished, one of the children of the root is selected as the best move. Final move selection techniques include choosing the max child (the child with the highest win rate), the robust child (the child with the highest visit count), the robust-max child (the child with both the highest win rate and visit count, where the search is continued until such a child exists), or the secure child (the child that maximizes a lower confidence bound).

It is straightforward to handle multiple-players in MCTS. The difference with the application to two-player games is that, after each playout is finished, instead of returning two reward values for each player, a tuple of $N$ values, is backpropagated in the tree. While more commonly used in perfect information turn based games such as Go or Chess, MCTS has also been extended to games with simultaneous moves \cite{lanctot2013monte,tak2014monte,perick2012comparison} and to imperfect information games \cite{cowling2012information}. MCTS based game playing combined with deep reinforcement learning has recently shown human level performance in highly complex games such as Go \cite{silver2016mastering}.

\subsection{Simultaneous Move MCTS}
One approach to simultaneous move MCTS is to serialize the game and apply standard MCTS \cite{balla2009uct}. In such a setting the search player chooses a move and the opponent is allowed to know which move the player chose and can respond accordingly. This leads to defensive play as the search player will learn to play the move that has the least chance of being penalized by an opponent. If this order is reversed, it would result in an aggressive play for the search player. Thus, serializing simultaneous move game produces artefact of defensive/offensive plays which may not be desirable depending on the application.

In order to properly account for simultaneous moves, we follow a strategy similar to the one proposed in \cite{tak2014monte} where we select both actions simultaneously and independently, and then update the state of the game. Since we treat both moves simultaneously, edges in the game tree are associated to pairs of moves $\act$ where $\actp$ denotes the move selected by player 1 and $\acto$ denotes the move selected by player 2. With this modification, while the expansion, simulation and backpropogation step remain the same as in sequential MCTS the selection step needs to be modified. Algorithm \ref{algo:SMCTS} provides a pseudocode description of the simultaneous move MCTS (SMCTS). The selection of a move is done in two steps (see BESTCHILD procedure): first, a set of statistics $\stat(\tnode)$ is extracted at each node $\tnode$ in the game tree, and then a selection policy $\selPol$ is invoked to choose the move given this information.

For a tree node $\tnode$, let $\actSet_i(\tnode(\cstatevec))=\{\actg_{i1},\actg_{i2},\cdots,\actg_{iM_i(\tnode)}\}$ be the available actions to the players $i=1,2$. For any $\actg\in \actSet_1(\tnode(\cstatevec))$, define
\begin{equation}\label{eq:C}
C_1(\tnode,\actg)=\{\mbox{child}(\tnode,\actg,\actg^\prime)|\actg^\prime\in \actSet_2(\tnode(\cstatevec))\},
\end{equation}
and, similarly for $C_2$. The statistics vector associated with player $i$ is defined as
\begin{equation}\label{eq:stat}
\stat_i(\tnode)=(\actg_{i1},q_{i1},n_{i1},\actg_{i2},q_{i2},n_{i2},\cdots,\actg_{iM_i(\tnode)},q_{iM_i(\tnode)},n_{iM_i(\tnode)})^*,
\end{equation}
where, $*$ denotes vector/matrix transpose, and
\begin{equation}\label{eq:rewmarg}
n_{ij}=\sum_{\tnode^\prime\in C_i(\tnode,\actg_{ij})} N(\tnode^\prime), \qquad q_{ij}=\frac{\sum_{\tnode^\prime\in C_i(\tnode,\actg_{ij})} \Qval(\tnode^\prime)}{n_{ij}},
\end{equation}
where, $\Qval_i(\tnode)$ is the accumulated rewards at node $\tnode$ for the players $i=1,2$ as described in the BACKUP procedure, see Algo. \ref{algo:SMCTS}.

Given the vector of statistics $\stat_i(\tnode)$, the selection policy $\selPol$ is an algorithm that selects an action $\actg_{ij^*}$ for player $i$ according to an index-based multi-arm bandit policy,
\begin{equation}\label{eq:selPol}
j^*=\selPol(\stat_i(\tnode))=\argmax_{j\in [1,M_i(\tnode)]} \mbox{index}(q_{ij},n_{ij},n_i),
\end{equation}
where, $n_i=\sum_{j=1}^{M_i(\tnode)}n_{ij}$. Several deterministic and stochastic $\mbox{index}$ policies have been explored in the literature \cite{perick2012comparison}, we discuss some examples below:
\begin{itemize}
  \item \textit{UCB1}: is a deterministic index policy motivated by UCT (see Eq. ~(\ref{eq:uct})),
  \begin{equation}\label{eq:UCB1}
    \mbox{index}(q_{ij},n_{ij},n_i)=q_{ij}+c\sqrt{\frac{\ln n_i}{n_{ij}}},
  \end{equation}
  where, $c$ is a parameters which enables the control of exploration/exploitation.
  \item \textit{Thompson Sampling}: is stochastic index policy, where first a random score is drawn from a $\mbox{beta}$ distribution
   \begin{equation}\label{eq:thom}
   sc_i(j)\sim\mbox{beta}(c_1+q_{ij}n_i,c_2+(1-q_{ij})n_{ij}), j\in [1,M_i(\tnode)],
  \end{equation}
  and then the move which maximizes this score is selected. The constants $c_1,c_2$ are two tunable parameters that reflect prior knowledge on reward expectations.
\end{itemize}
Some other index functions e.g. \textit{UCB1-Tuned}, \textit{UCB-V} and \textit{PBBM} also require one to keep track of empirical standard deviation of the rewards at each node, see \cite{perick2012comparison} for details and other choices of index function.

\begin{algorithm}
\caption{Simultaneous Move MCTS (SMCTS)}\label{algo:SMCTS}
\begin{algorithmic}[1]

\Procedure{UCTSEARCH}{$\cstatevec_0$}
\State Create root node $\tnode_0$ with state $\cstatevec_0$
\While{within computational budget} \label{algo:budget}
\State $\tnode_l\leftarrow$ SELECTION$(\tnode_0)$
\State $\Delta_1,\Delta_2 \leftarrow$ SIMULATE$(\cstatevec(\tnode_l))$
\State BACKUP$(\tnode_l,\Delta_1,\Delta_2)$
\EndWhile
\State \textbf{return} $b$
\EndProcedure
\\
\Procedure{SELECTION}{$\tnode$}
\While{$\tnode$ is nonterminal}
\If{$\tnode$ not fully expanded}
\State \textbf{return} EXPAND$(\tnode)$
\Else
\State $\tnode\leftarrow$ BESTCHILD$(\tnode)$
\EndIf
\EndWhile
\State \textbf{return} $\tnode$
\EndProcedure
\\

\Procedure{EXPAND}{$\tnode$}
\State choose $\act \in$ untried action from $\actSet(\cstatevec(\tnode))$
\State $\cstatevec(\tnode^\prime)=\trans(\cstatevec(\tnode),\act)$
\State add the new child $\tnode^\prime$ to $\tnode$
\State \textbf{return} $\tnode^\prime$
\EndProcedure
\\

\Procedure{BESTCHILD}{$\tnode$}
\State $\actp=\pi(\stat_1(\tnode))$
\State $\acto=\pi(\stat_2(\tnode))$
\State \textbf{return} child$(\tnode,\actp,\acto)$
\EndProcedure
\\

\Procedure{SIMULATE}{$\cstatevec$}
\While{$\cstatevec$ is nonterminal} \label{algo:termcond}
\State choose $\act\in\actSet(\cstatevec)$ uniformly at random
\State $\cstatevec\leftarrow\trans(\cstatevec,\act)$ \label{algo:randro}
\EndWhile
\State \textbf{return} rewards $\utl_1(\cstatevec),\utl_2(\cstatevec)$ for state $\cstatevec$
\EndProcedure
\\

\Procedure{BACKUP}{$\tnode$,$\Delta_1,\Delta_2$}
\While{$\tnode$ is not null}
\State $N(\tnode)\leftarrow N(\tnode)+1$
\State $\Qval_1(v)\leftarrow\Qval_1(v)+\Delta_1 $
\State $\Qval_2(v)\leftarrow\Qval_2(v)+\Delta_2 $
\State $\tnode\leftarrow$ parent of $\tnode$
\EndWhile
\EndProcedure

\end{algorithmic}
\end{algorithm}

\section{Problem Formulation}\label{sec:prob}
We consider a relative maneuvering problem between two aircrafts $i=1,2$, whose
dynamics in discrete time is governed by
\begin{equation}\label{eq:dyn}
\statevec_i(k+1)=\mathbf{\dyn}(\statevec_i(k),\ctrvec_i(k);\paramvec_i),
\end{equation}
where, $\statevec_i(k)=(\pos_i(k),\velvec_i(k),\att_i(k),\omgvec_i(k))^*\in \cSet\subset \R^{\ns}$ is the state vector at time instant $k$ with $\pos_i(k)$ denoting the aircraft position vector, $\att_i(k)$ is its attitude vector, and $\velvec_i(k)$ and $\omgvec_i(k)$ are its linear velocity and attitude rate vectors, respectively. The vector $\ctrvec_i(k) \in \ctrlset \subset \R^{\nc}$ is the control vector, and $\paramvec_i$ are the aircraft parameters related to bounds on maximum speed, turn rate, etc.  which dictate its maneuverability. We parameterize the control inputs in form of a finite set of basic maneuvers $\actSet$, such that any $\actg\in\actSet$ will correspond to predetermined choice of control input vector $\ctrvec^\actg$. Thus, motion of each aircraft can be controlled by choosing a sequence of basic maneuvers from the sets $\actSet_i,i=1,2$, respectively.

\begin{figure}
  \begin{center}
  \includegraphics[trim=0cm 7cm 0cm 0cm,clip,scale=0.5]{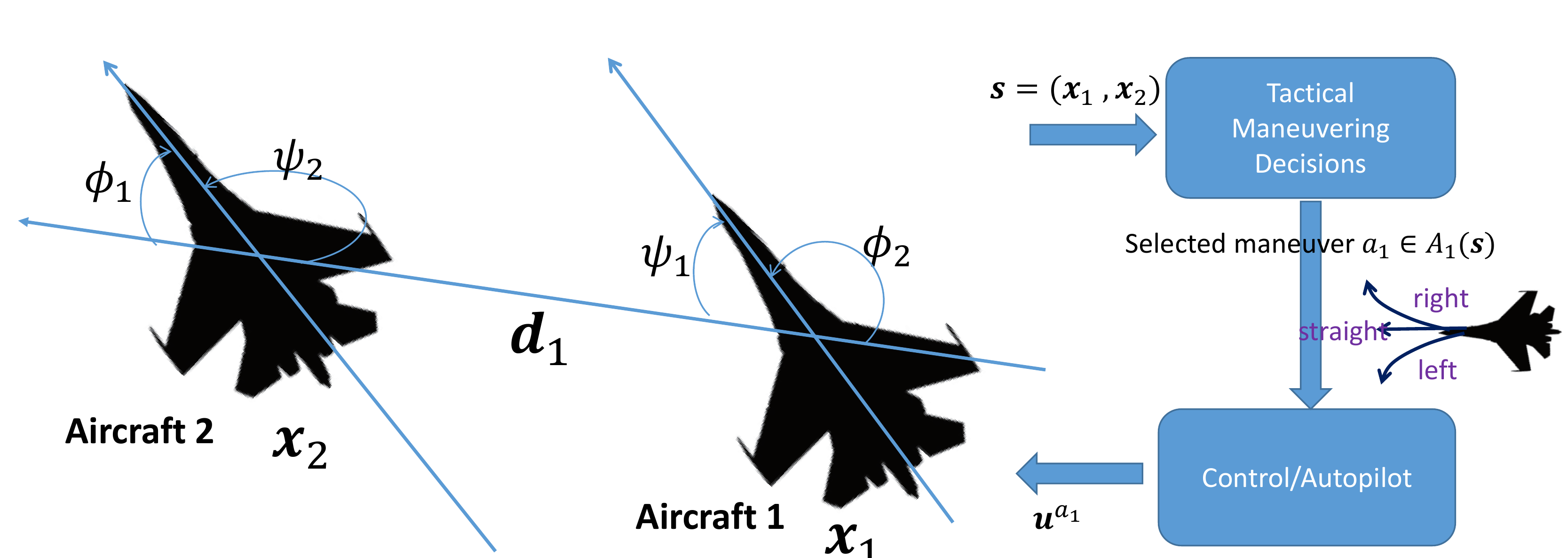}
  \caption{Schematic of relative configuration between two aircrafts.}\label{fig:relstate}
  \end{center}
\end{figure}

In order to characterize the relative configuration of the two aircrafts at time instant $k$, we introduce a vector $\rstatevec_i(k)=(\brg_i(k),\angoff_i(k),\dist_i(k))^*$ (see Fig. \ref{fig:relstate}), where, for $i=1$
\begin{eqnarray}
  \brg_1(k)&=& \arccos(\frac{\distvec_1(k)\cdot\velvec_1(k)}{||\distvec_1(k)|| ||\velvec_1(k)||} ), \qquad \angoff_1(k) = \arccos(\frac{\distvec_1(k)\cdot\velvec_2(k)}{||\distvec_1(k)|| ||\velvec_2(k)||} ),\quad \dist_1(k)=||\distvec_1(k)||,
\end{eqnarray}
with $\distvec_1(k)=\pos_2(k)-\pos_1(k)$ being the line-of-sight vector of aircraft $i=2$ relative to aircraft $i=1$. Similar definitions apply for aircraft $i=2$. Note that by definition,
\begin{equation}
  \brg_1(k)+\angoff_2(k) = \pi, \qquad \brg_2(k)+\angoff_1(k) = \pi. \label{eq:angcnst}
\end{equation}
The terminal set for $i-$th aircraft is defined as
\begin{equation}\label{eq:termset}
\cSetT_i=\{(\statevec_1,\statevec_2)|\dist_{min}<\dist_i<\dist_{max},\quad |\angoff_i|<\angoff_{\mbox{max}}  \quad \quad |\brg_i|<\brg_{\mbox{max}}\},
\end{equation}
where, $\dist_{min},\dist_{max},\angoff_{\mbox{max}},\brg_{\mbox{max}}$ are pre-defined parameters. The goal for each aircraft is to reach it's terminal state $\cSetT_i$ before the other aircraft can do the same. This naturally leads to a game theoretic setting.

\subsection{Game Representation}\label{sec:relMgame}
The following provides a game representation of the relative maneuvering problem:
\begin{itemize}
\item \textbf{State}: The game state is defined by the combined states of the two aircrafts $\cstatevec=(\statevec_1,\statevec_2)$ with the terminal set for the game being $\cSetT=\cSetT_1\bigcup\cSetT_2$.
\item \textbf{Action}: $\actSet = \actSet_1 \times \actSet_2$, where $\actSet_i$ is set of maneuvers available to the $i-$th aircraft.
\item \textbf{Transition function}: $\trans(\cstatevec,\act)=\left(\begin{array}{c}
                                                           \mathbf{\dyn}(\statevec_1,\actp;\paramvec_1)  \\
                                                           \mathbf{\dyn}(\statevec_2,\acto;\paramvec_2)
                                                          \end{array}\right)
$, where for brevity
\begin{equation}\label{eq:dynman}
\mathbf{\dyn}(\statevec,\actg;\paramvec)\equiv\mathbf{\dyn}(\statevec,\ctrvec^\actg;\paramvec)
\end{equation}
\item \textbf{Reward}: The objective of reaching the terminal state can be captured in the reward functions defined as follows:
\begin{equation}\label{eq:termreward}
  \utl_1(\cstatevec)=
   \begin{cases}
            1 & \mbox{if} \quad \cstatevec\in \cSetT_1 \\
            -1 & \mbox{if} \quad \cstatevec\in \cSetT_2 \\
            0 & \mbox{otherwise}
        \end{cases}.
\end{equation}
and similarly for $\utl_2$. Thus, $\utl_1(\cstatevec)+\utl_2(\cstatevec)=0$ and so the game is zero sum.

\end{itemize}

\section{Algorithm}\label{sec:algo}
In order to solve the relative maneuvering game introduced in the Section ~\ref{sec:relMgame}, we use the SMCTS Algo. \ref{algo:SMCTS} but with several modifications which are discussed below.

\paragraph{Modified Playout Step:} Since during simulation step reaching a terminal state could typically require many steps, in order to reduce the computational time we introduce a fixed simulation horizon $\thor_{\mbox{sim}}$. Consequently, line number ~(\ref{algo:termcond}) in Algo. \ref{algo:SMCTS} is modified so that the simulation step is terminated if a terminal state is reached or the number of time steps reaches $\thor_{\mbox{sim}}$.
With this approach since the simulation could end in a non-terminal state, we need a mechanism to
estimate the value of this state. Intuitively, if this non-terminal state is closer to the terminal state, this state should be a better state if there is no other information. To capture this we use reward shaping from \cite{mcgrew2010air}:
\begin{equation}\label{eq:rewshapetotal}
\utl^s_i(\cstatevec)=w\utl_i(\cstatevec)+(1-w)\overline{\utl}_i(\cstatevec),
\end{equation}
where, $\utl_i$ is the terminal reward as given in Eq. ~(\ref{eq:termreward}),
\begin{eqnarray}\label{eq:rewshape}
  \overline{\utl}_i(\cstatevec) &=&\frac{1}{2}-\frac{1}{2}[1-(1-\angoff_i/\pi)-(1-\brg_i/\pi)][\exp(-\frac{|\dist_i-\dist_{nom}|}{r_d})]
\end{eqnarray}
is the reward shaping term with parameters $\dist_{nom}$ and $r_d$, and $w\in(0,1)$ is a weight which determines relative importance of $\utl_i$ and $\overline{\utl}_i$. Finally, we aggregate the total reward over the horizon $\thor_{\mbox{sim}}$ in a discounted fashion, i.e.
\begin{equation}
\tilde{\utl}_i(\cstatevec)=\sum_{k=0}^{\thor_{\mbox{sim}}-1}\disfac^{k}\utl^s_i(\cstatevec(k)), \label{eq:totalrew}
\end{equation}
where, $\cstatevec(k),k=0,\cdots,\thor_{\mbox{sim}}$ are the game states encountered during playout, and $\disfac\in (0,1]$ is a discount factor.

\paragraph{Playout Policy:} During simulation step in addition to the random playout (see line number ~(\ref{algo:randro}) in Algo. \ref{algo:SMCTS})) we consider two additional  options. Let $\cstatevec=(\statevec_1,\statevec_2)$ be a game state during simulation:
\begin{itemize}
  \item Greedy playout: In this approach player $1$ chooses action $\actg_{1j^*}$, where
\begin{equation*}
  j^*=\argmax_{j} \utl^s_1(\cstatevec_j^\prime)
\end{equation*}
  where, $\cstatevec_j^\prime=(\dyn(\statevec_1,\actg_{1j};\paramvec_1),\statevec_2)$  is the transition game state assuming other player 2 is static, and similarly for player 2.

  \item Matrix game playout: Let $\payoff_i=[\payoffelem^i_{jk}]$ be 1-step payoff matrix w.r.t to player $i$ where $\payoffelem^i_{jk}=\utl^s_i(\trans(\cstatevec,\actg_{1j},\actg_{2k}))$ is the reward at next state resulting from actions $\actg_{1j}\in\actSet_1(\cstatevec),\actg_{2k}\in\actSet_2(\cstatevec)$. Then the player chooses action $\actg_{ij^*}$ according to a mixed strategy such that $j^*\sim \probvec_i$, where $\probvec_i=(\probelem_{i1},\cdots,\probelem_{i|\actSet_i(\cstatevec)|})^*$ is probability vector over set $\{1,\cdots,|\actSet_i(\cstatevec)|\}$ obtained by solving (via linear program) a max-min optimization
\begin{eqnarray}
   && \max_{\probvec_i} \min_{j=1,\cdots,|\actSet_{3-i}(\cstatevec)|}(\payoff_i^*\probvec_i)_j\notag \\
  && \mbox{subject to}\quad \probvec_i\geq 0,\quad \sum_{j=1}^{|\actSet_i(\cstatevec)|}\probelem_{ij}=1, \label{eq:matgame}
\end{eqnarray}
where, $(\payoff_i^*\probvec_i)_j$ denotes the $j-$th element of vector $\payoff_i^*\probvec_i$.
\end{itemize}
We will refer to SMCTS with the greedy playout and the matrix game playout as SMCTS-G and SMCTS-M, respectively. Finally, to control computational budget in SMCTS (see line \ref{algo:budget} in Algo. \ref{algo:SMCTS}), we place a bound on total allowed size of search tree denoted by $M_{\mbox{tree}}$, where tree size is the total number of nodes in the tree.

\section{Numerical Studies}\label{sec:sim}
For numerical studies we consider relative maneuvering problem in 2d with the aircraft dynamics governed by Eq. ~(\ref{eq:dyn}) and $\dyn(\statevec,\actg,\paramvec)$ as described in Algo. \ref{algo:dyn}. In this 2d representation, $\statevec=(x,y,v,\theta,\zeta)^*$, $\actg\in\actSet=\{\mbox{left,straight,right}\}$ are the basic maneuvers, and $\paramvec=(v,\dot{\zeta},\zeta_{max},\Delta t,N_s)$. The two aircrafts are assumed to be identical, and the parameters used in simulation are listed in the Table ~\ref{tbl:param} along with the reward parameters appearing in Eq. ~(\ref{eq:termset}) and Eq. ~(\ref{eq:rewshape}).

\begin{algorithm}
\caption{Aircraft Dynamics}\label{algo:dyn}
\begin{algorithmic}[1]
\Procedure{$\dyn$}{$\statevec,\actg,\paramvec$}
\State $k=0$
\While{$k\leq N_s$}
\If{$\actg=$ left}
\State $\zeta=\max(\zeta-\dot{\zeta}\Delta t,-\zeta_{max})$
\ElsIf{$\actg=$ right}
\State $\zeta=\min(\zeta+\dot{\zeta}\Delta t,\zeta_{max})$
\EndIf
\State $\theta=\theta+\frac{g}{v}\tan\zeta \Delta t$
\State $x=x+v\cos \theta \Delta t$
\State $y=y+v\sin \theta \Delta t$
\State $k=k+1$
\EndWhile
\EndProcedure
\end{algorithmic}
\end{algorithm}

\begin{table}[hbt!]
\centering
\begin{tabular}{|c|c|c|c|c|c|c|c|c|c|}
  \hline
  Aircraft Id & $v$ & $\dot{\zeta}$ & $\zeta_{max}$ & $\dist_{min}$ & $\dist_{max}$ & $\dist_{nom}$ & $r_d$ & $\brg_{max}$ & $\angoff_{max}$ \\ \hline
  $i=1,2$ & $2.5$m/s & $45^\circ$/s & $23^\circ$& $0.1$m & $3$m  & $2$m  & $18$m & $30^\circ$& $60^\circ$ \\
  \hline
\end{tabular}
\caption{Aircraft and reward parameters.}\label{tbl:param}
\end{table}

We will next compare maneuvering performance of the two aircrafts based on different tactics listed in the Table \ref{tbl:cases}. We also consider variations in aircraft parameters to study the impact of physical differences in maneuverability. The MG (matrix game) approach listed in the table refers to the mixed strategy obtained by solving Eq. (\ref{eq:matgame}) based on one-step payoff matrix; note no MCTS is used here. The pseudo code for game playing is shown in Algo. ~\ref{algo:simgame}, where, $\cstatevec$ is the starting game state, and $\thor_{\mbox{game}}$ is the time horizon over which game is played.  By Player1Tactics/Player2Tactics we denote the maneuvering strategy each aircraft selects based on the Table \ref{tbl:cases}.

\begin{table}[hbt!]
\centering
\begin{tabular}{|c|c|c|c|c|}
  \hline
   Case No.&  Aircraft 1 & Aircraft 2  & Aircraft 1 $\paramvec_1$ & Aircraft 2 $\paramvec_2$\\ \hline
   I & MG & MG & Same as in Table \ref{tbl:param} & Same as in Table \ref{tbl:param}
   \\ \hline
   II & SMCTS-M & MG & Same as in Table \ref{tbl:param} & Same as in Table \ref{tbl:param}
   \\ \hline
   III & SMCTS-M & MG & Same as in Table \ref{tbl:param} & $\dot{\zeta}=22.5^\circ$/s, others same as in Table \ref{tbl:param}
   \\ \hline
   IV & SMCTS-M & MG & $\dot{\zeta}=22.5^\circ$/s, others same as in Table \ref{tbl:param} & Same as in Table \ref{tbl:param}
   \\ \hline
\end{tabular}
\caption{Different cases representing different approaches for generating maneuvering tactics.}\label{tbl:cases}
\end{table}

\begin{algorithm}
\caption{Game Simulation}\label{algo:simgame}
\begin{algorithmic}[1]
\Procedure{SIMGame}{$\cstatevec$}
\State $k=0$
\While{$k\leq \thor_{\mbox{game}}$ or $\cstatevec\in \cSetT$} \label{algo:budget}
\State $\actg_1=$ Player1Tactics $(\cstatevec)$
\State $\actg_2=$ Player2Tactics $(\cstatevec)$
\State $\cstatevec=\trans(\cstatevec,\actg_1,\actg_2)$
\State $k\leftarrow k+1$
\EndWhile
\State \textbf{return} game outcome
\EndProcedure
\end{algorithmic}
\end{algorithm}

For each case listed in the table we compare the performance both qualitatively and quantitatively. For qualitative comparison we visualize the trajectories of two aircrafts starting from specific positions.
For quantitative comparison we use a Monte Carlo (MC) study, where we randomly initialize the state of the two aircrafts, simulate the game for a fixed horizon $\thor_{\mbox{game}}$, and record the final game outcome as to whether aircraft 1 wins, loses or the game is a draw.  Note that we symmetrize the initial conditions, i.e. for each initial condition $\cstatevec=(\statevec_1,\statevec_2)$ we also consider $\cstatevec^\prime=(\statevec_2,\statevec_1)$ where aircrafts switch their starting state from which the game begins. This process is repeated for a total of $M_s$ times, and let $M_{w1}$ be total of trials in which aircraft 1 wins, $M_{w2}$ be total of trials in which aircraft 2 wins, and $M_d$ be the total number of trails in which the game is a draw. This game statistics can be summarized in form of the first aircraft’s probability of win $\prob_{w1} = M_{w1}/M_s$, second aircraft’s probability of win $\prob_{w2} = M_{w2}/M_s$ and the probability of the draw $\prob_d = M_d/M_s$.

\subsection{Preliminary Results}
For the simulations, we used $\Delta t=0.05$sec and $N_s=20$ so that each basic maneuver is held for $1$ sec, $\thor_{\mbox{game}}=70$ sec and $\thor_{\mbox{sim}}=10$ sec.  Furthermore, we use $c=0.2$ in the UCB1 formula (see Eq. ~(\ref{eq:UCB1})), $w=0.5$ for reward shaping in Eq. ~(\ref{eq:rewshapetotal}) and $\disfac=0.8$ in Eq. ~(\ref{eq:totalrew}) for the total simulation reward. A total of $M_s=100$ MC trails were used for all the cases considered.

\begin{figure}
  \begin{center}
  \includegraphics[trim=0cm 8cm 0cm 8cm,clip,scale=0.7]{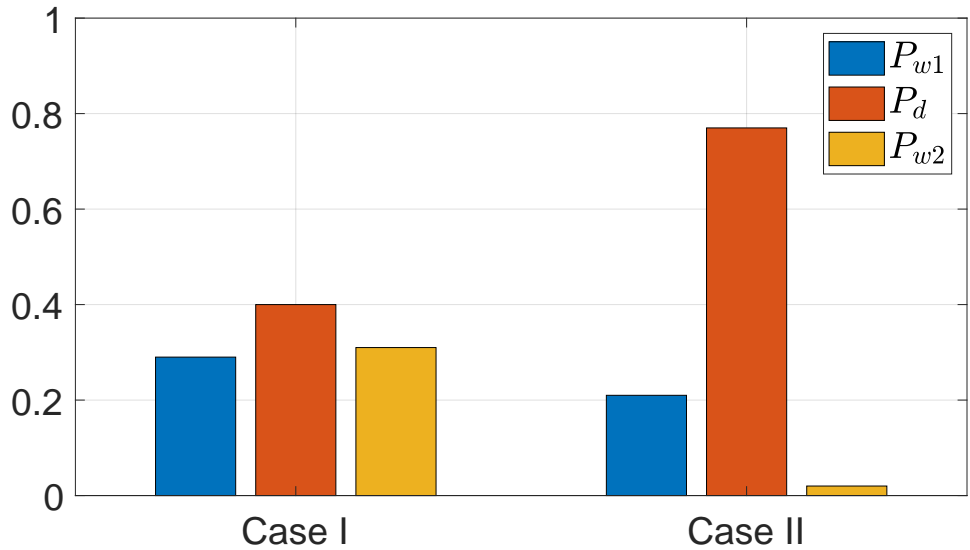}
  \caption{Results of Monte Carlo trials, comparing $\prob_{w1}$ $\prob_{w2}$ and $\prob_d$ for Case I and Case II listed in Table \ref{tbl:cases}.}\label{fig:hist}
  \end{center}
\end{figure}

\begin{figure}
 \begin{center}
  \includegraphics[trim=0cm 5.5cm 0cm 0cm,clip,scale=0.5]{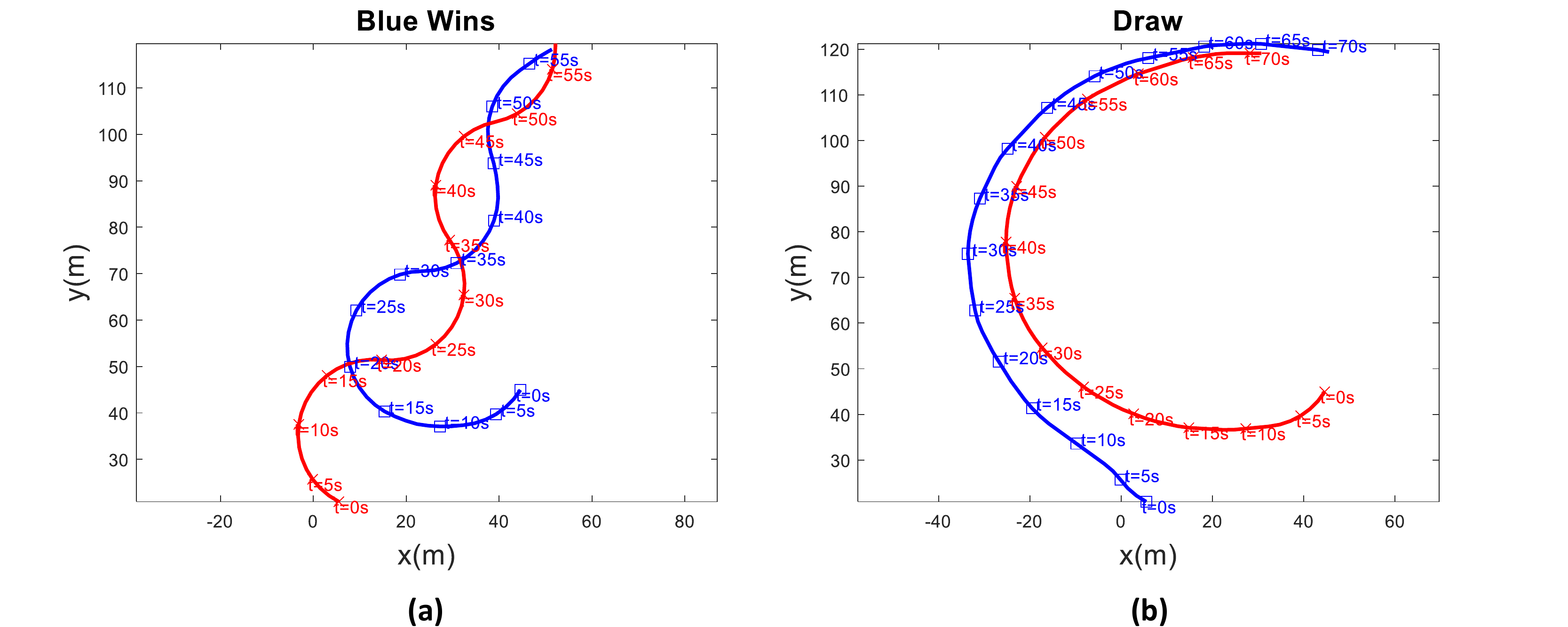}
  \caption{A trail from MC for case II, aircraft 1 is shown in blue track and aircraft 2 is shown in red track along with time stamps. a) aircraft 1 starts in an advantageous position and wins , and b)  aircraft 2 starts in an advantageous position, however aircraft 1 is able to evade.}\label{fig:case12}
  \end{center}
\end{figure}

Figure ~\ref{fig:hist} shows the  $\prob_{w1} $ $\prob_{w2}$ and $\prob_d$ for cases I and II listed in the Table \ref{tbl:cases}. Given aircrafts have same parameters and hence same physical maneuvering limits, for case I we see similar win rates for both the aircrafts. However, for case II aircraft 1 uses SMCTS-M (we used $M_{\mbox{tree}}=9$), and with longer term planning is able to evade situations in which aircraft 2 starts in advantageous positions, and thus converting them into draws. At the same time aircraft 1 win probability $\prob_{w1}$ is reduced slightly compared to case I. It appears that SMCTS-M results in more conservative behavior compared to greedy 1-step matrix game approach. In fact, SMCTS-M is based on self-play so aircraft 1 assumes aircraft 2 is also playing the best response. However, this assumption is not true as aircraft 2 plays using greedy 1-step matrix game approach, and thus could to more conservative outcomes for aircraft 1. To illustrate this, an example MC trial for case II is shown in Fig. ~\ref{fig:case12} with aircraft 1 shown in a blue track and aircraft 2 in a red track. Also shown are the time stamps along each track.  In  Fig. ~\ref{fig:case12}a  aircraft 1 starts in an advantageous position and wins. In Fig. ~\ref{fig:case12}b states of aircraft 1 and 2 are reversed, and so aircraft 2 starts in same relative advantageous position, however aircraft 1 is able to evade and the game draws.

\begin{figure}
  \begin{center}
  \includegraphics[trim=0cm 8cm 0cm 8cm,clip,scale=0.6]{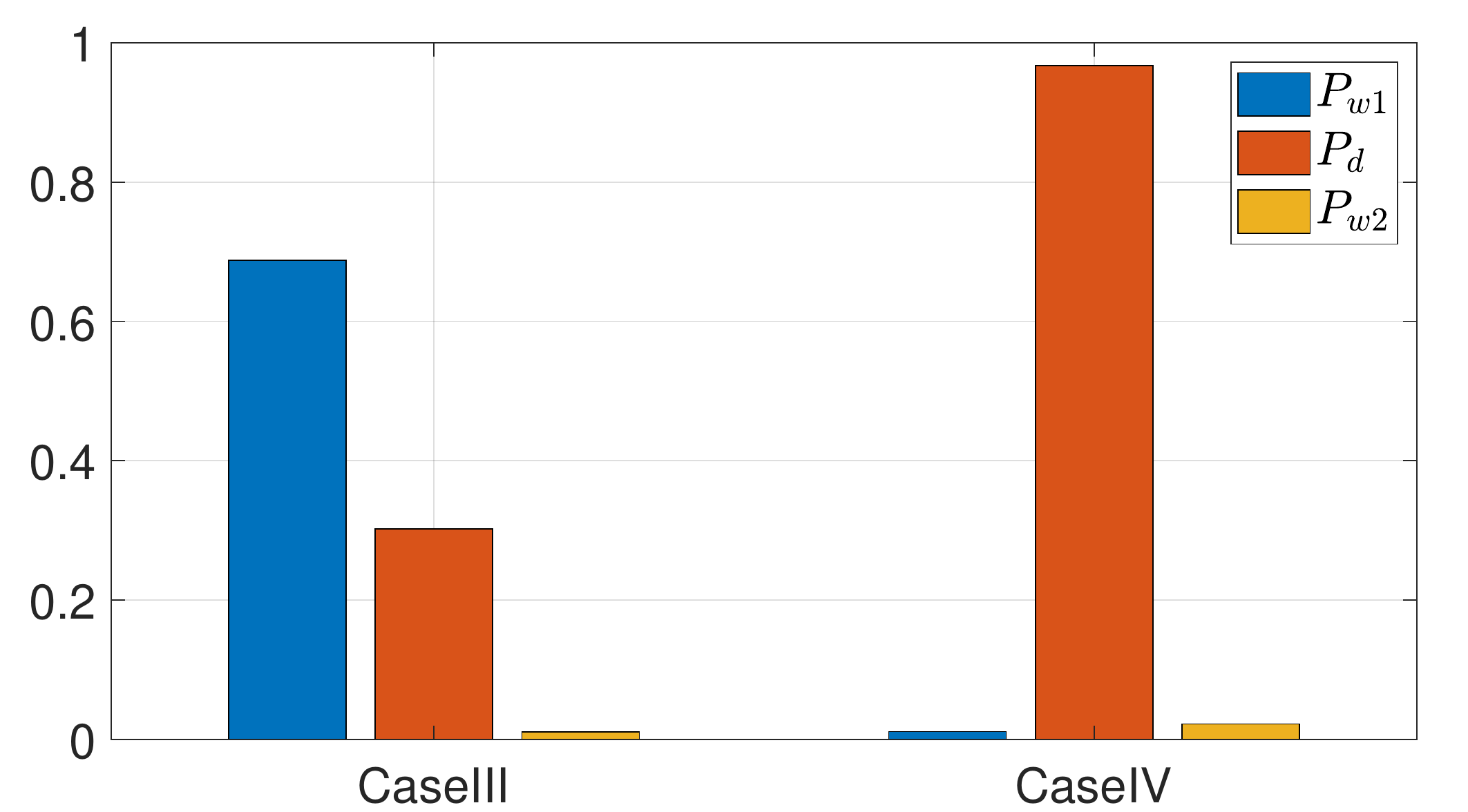}
  \caption{Results of Monte Carlo trials, comparing $\prob_{w1}$ $\prob_{w2}$ and $\prob_d$ for Case III and Case IV listed in Table \ref{tbl:cases}.}\label{fig:hist34}
  \end{center}
\end{figure}

Figure ~\ref{fig:hist34} shows the  $\prob_{w1} $ $\prob_{w2}$ and $\prob_d$ for cases III and IV listed in the Table \ref{tbl:cases}. For case III, blue is twice as more maneuverable than red while for case IV red is twice as more maneuverable than blue as dictated by choice of $\dot{\zeta}$. For case III, with superior manueverablity and superior tactics (SMCTS-M ) blue is able to win with a $70\%$ rate. In case IV, despite red having superior manueverablity, blue due to its superior tactics is able to convert loses into draws as was seen in Case II. 

\section{Conclusion} \label{sec:conc}
In this paper we applied the simultaneous move MCTS for online solution of tactical maneuvering between two aircrafts. In particular we performed numerical studies exploring different variations of the SMCTS algorithm including modifications to the simulation and playout step. These preliminary studies demonstrate feasibility/utility of the proposed framework.

While MCTS based online game solving offers an anytime solution approach which can adapted on the fly given the context (e.g. changing aircraft dynamic capabilities, accounting for opponent aircraft tactics etc.), it can still pose a challenge from real time implementation perspective when decisions need to be made at very fast time scales. Thus, it
would be worthwhile to explore approaches to accelerate MCTS, for example hardware acceleration using parallelization \cite{browne2012survey}, ability to reuse computations from past tree searches for the current search \cite{powley2014information}, and using learning methods (e.g. reinforcement learning) to off-line learn to estimate value of game states and use that information online to speedup tree search \cite{silver2016mastering}. Further practical considerations would require accounting for imperfect and/or partial aircraft state information and ability to adapt to  and exploit the opponent's behavior. Extensions to teams of multiple aircrafts is another important avenue for future research.

\section*{Appendix}
Funding provided by Raytheon Technologies Research Center is greatly appreciated.

\section*{Acknowledgments}

\bibliography{references}

\end{document}